\documentclass[10pt,twocolumn,letterpaper]{article}

\usepackage{cvpr} 
\usepackage{amsfonts} 
\usepackage[accsupp]{axessibility}
\usepackage{dblfloatfix}
\usepackage{nicefrac}       
\usepackage{microtype}      
\usepackage{graphicx}
\usepackage{array}
\usepackage{amsmath}
\usepackage{amssymb}
\usepackage{pifont}
\usepackage{booktabs}
\usepackage{times}
\usepackage{graphicx}
\usepackage{amsmath}
\usepackage{amssymb}
\usepackage{dsfont}
\usepackage{siunitx}
\usepackage{multirow}
\usepackage{enumitem}
\usepackage{tablefootnote}
\usepackage{bbm}
\usepackage{xspace}
\usepackage{mathtools}
\usepackage[normalem]{ulem}
\usepackage{tabularx}
\usepackage{wrapfig}
\usepackage{flushend}
\usepackage{outlines}
\usepackage{bbding}

\usepackage{url}
\usepackage{graphicx}
\usepackage{multirow}
\usepackage{caption}
\usepackage{array}
\usepackage{arydshln}
\usepackage{siunitx}
\usepackage{pifont}
\usepackage[pro]{fontawesome5}
\usepackage{listings}
\usepackage{verbatim}
\usepackage{amsmath}
\usepackage{overpic}
\usepackage{url}
\usepackage{multirow}
\usepackage{amsmath}
\usepackage{amssymb}
\usepackage{pifont}
\usepackage{float}
\usepackage{tikz}
\usepackage{bbding}
\usepackage{media9}
\usepackage{wrapfig}
\usepackage{animate}
\definecolor{mygray}{HTML}{f0f0f0}
\definecolor{mygreen}{HTML}{35cd2d}
\usepackage{tabulary,multirow,overpic}
\usepackage{arydshln}

\usepackage[hang,flushmargin]{footmisc} 

\usepackage[dvipsnames]{xcolor}
\usepackage[inkscapelatex=false]{svg}

\definecolor{cvprblue}{rgb}{0.21,0.49,0.74}
\definecolor{dgreen}{rgb}{0.0,0.6,0.0}
\usepackage[pagebackref,breaklinks,colorlinks,citecolor=cvprblue]{hyperref}

\def\paperID{3130} 
\def\confName{CVPR}
\def\confYear{2025}

\title{Vinci: A Real-time Embodied Smart Assistant based on Egocentric Vision-Language Model}

\author{\fontsize{11}{15}\selectfont Yifei Huang$^1$, Jilan Xu$^1$, Baoqi Pei$^1$, Yuping He$^1$, Guo Chen$^1$, Lijin Yang$^1$, Xinyuan Chen$^1$, Yaohui Wang$^1$ \\ \fontsize{11}{15}\selectfont Zheng Nie$^2$, Jinyao Liu$^2$, Guoshun Fan$^2$, Dechen Lin$^2$, Fang Fang$^2$, Kunpeng Li$^2$, Chang Yuan$^2$\\ \fontsize{11}{15}\selectfont Yali Wang$^3$, Yu Qiao$^3$, Limin Wang$^3$ \\ \normalsize{\texttt{opengvlab@gmail.com} ~ Shanghai Artificial Intelligence Laboratory}
}

\begin{document}

\maketitle
\let\thefootnote\relax\footnotetext{ $^1$Model contributors $^2$ System contributors $^3$Corresponding authors
}
\begin{abstract}
We introduce Vinci, a real-time embodied smart assistant built upon an egocentric vision-language model. Designed for deployment on portable devices such as smartphones and wearable cameras, Vinci operates in an ``always on" mode, continuously observing the environment to deliver seamless interaction and assistance. Users can wake up the system and engage in natural conversations to ask questions or seek assistance, with responses delivered through audio for hands-free convenience. With its ability to process long video streams in real-time, Vinci can answer user queries about current observations and historical context while also providing task planning based on past interactions. To further enhance usability, Vinci integrates a video generation module that creates step-by-step visual demonstrations for tasks that require detailed guidance. We hope that Vinci can establish a robust framework for portable, real-time egocentric AI systems, empowering users with contextual and actionable insights. We release the complete implementation for the development of the device in conjunction with a demo web platform to test uploaded videos at \url{https://github.com/OpenGVLab/vinci}.

\end{abstract}
    
\vspace{-1em}
\section{Introduction}
\label{sec:intro}

In recent years, advancements in vision-language models (VLMs)~\cite{gpt4o,gpt4v,gemini,internvl2_5,QwenVL,li2023videochat,internvideo2} have unlocked new possibilities for creating intelligent systems capable of understanding and interacting with the visual world. Among these innovations, egocentric vision has gained significant attention due to its ability to capture the world from a user’s point of view~\cite{bandini2020analysis,plizzari2023outlook,damen2018scaling,egoexo4d,huang2020improving}. This perspective is particularly well-suited for applications on wearable devices and portable systems, where continuous observation and interaction can provide users with real-time, context-aware assistance~\cite{park2008wearable,bachlin2009swimmaster,danry2020wearable}.

\begin{figure*}
    \centering
    \includegraphics[width=\linewidth]{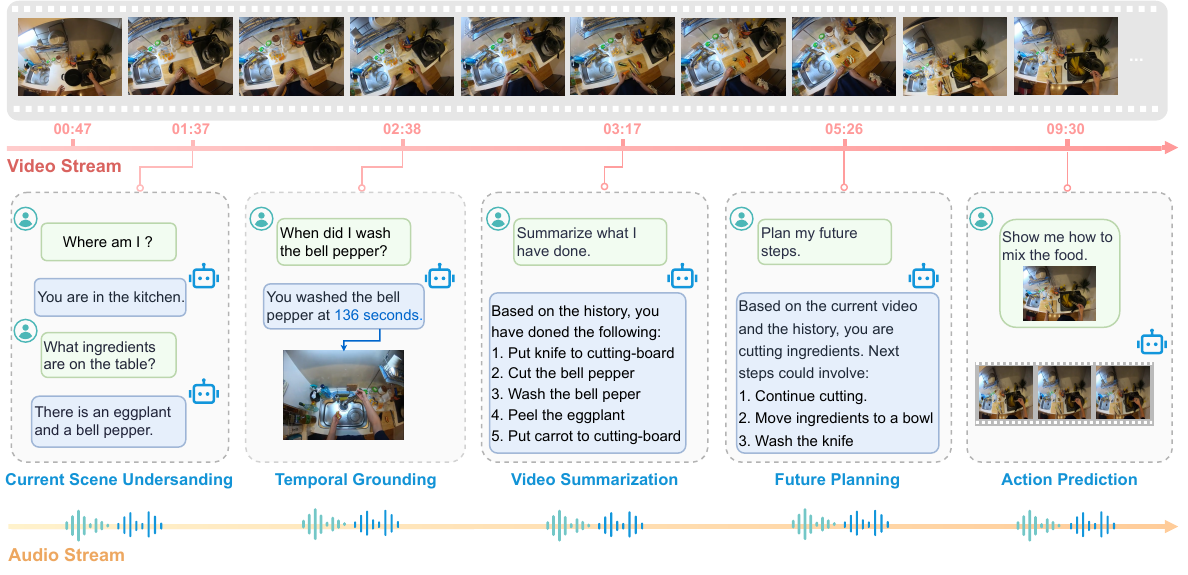}
    \caption{Overview of Vinci’s capabilities demonstrated through a streaming video timeline. At different timestamps, Vinci showcases its diverse real-time abilities: (1) Current scene understanding—providing detailed analysis of the ongoing activity and environment; (2) Temporal grounding—retrieving and referencing past events based on user queries; (3) Video summarization—offering concise summaries of key actions over time; (4) Future planning—predicting upcoming steps or actions based on historical context and current observations; and (5) Action prediction—generating a visual demonstration of the next likely action to assist users in task completion.}
    \label{fig:intro}
\end{figure*}

Despite the promise of egocentric vision-language systems~\cite{zhao2023learning,zhao2023training,pei2024egovideo,dou2024unlocking}, several challenges remain. Many models are not optimized for real-time processing, limiting their ability to operate continuously on resource-constrained devices such as smartphones or wearable cameras~\cite{park2020deep,janaka2024towards,janaka2024demonstrating,danry2020wearable}. Furthermore, these systems often lack the ability to retain and utilize historical context, making it difficult to answer questions or provide planning based on historical information~\cite{fathi2011learning,huang2018predicting,kazakos2019epic,li2023videochat,yang2022interact}. Also, current solutions lack the ability to bridge the gap between high-level task descriptions and actionable guidance, particularly when users require detailed, visible demonstrations~\cite{tekin2019h+,xu2024retrieval}.

To address these limitations, we introduce a new system termed \textbf{Vinci}, a real-time embodied smart assistant built on EgoVideo-VL, a newly trained egocentric vision-language model. Vinci is designed to deliver seamless assistance on portable devices, combining continuous observation with audio-based interaction. Users can activate the system with a voice command and engage naturally by asking questions, and responses are provided in spoken languages for hands-free convenience. Vinci processes long video streams in real-time, allowing it to answer questions about both the present and the past while enabling task planning based on historical context. Additionally, Vinci integrates a video generation module that produces visual how-to demonstrations, empowering users to tackle tasks requiring detailed visual guidance.

Vinci is composed of multiple integrated modules designed to deliver real-time performance and seamless user interaction. First, live streaming video and audio from portable devices is processed by an input module that transcribes the audio and pairs it with video streams. Through a backend server, these data are sent to model services, where the core model, EgoVideo-VL, generates responses by integrating current inputs, historical context, and user queries. The memory module, integrated with the EgoVideo-VL model, periodically records video timestamps and textual descriptions, even in the absence of user queries. This ensures a comprehensive log of user activities and environmental events. Vinci also features a video generation module that creates step-by-step how-to demonstrations in response to specific queries, offering users actionable guidance. Communication between the frontend devices and the backend server is facilitated via HTTP services, ensuring efficient and reliable data exchange. Together, these components empower Vinci to provide robust and context-aware assistance in dynamic, real-world scenarios.

Vinci embodies key innovations in real-time, egocentric AI systems, offering seamless audio-based interaction, real-time processing of live-streaming video and audio, and integration of historical context for robust responses. Its memory module ensures detailed activity logs, while the video generation and video retrieval modules provide visual how-to demonstrations for user guidance. To accelerate future research and development, we have released the model parameters alongside the full inference and deployment source code. We hope Vinci can inspire new advancements and applications in the field of egocentric AI systems.

\begin{figure*}
    \centering
    \includegraphics[width=0.8\linewidth]{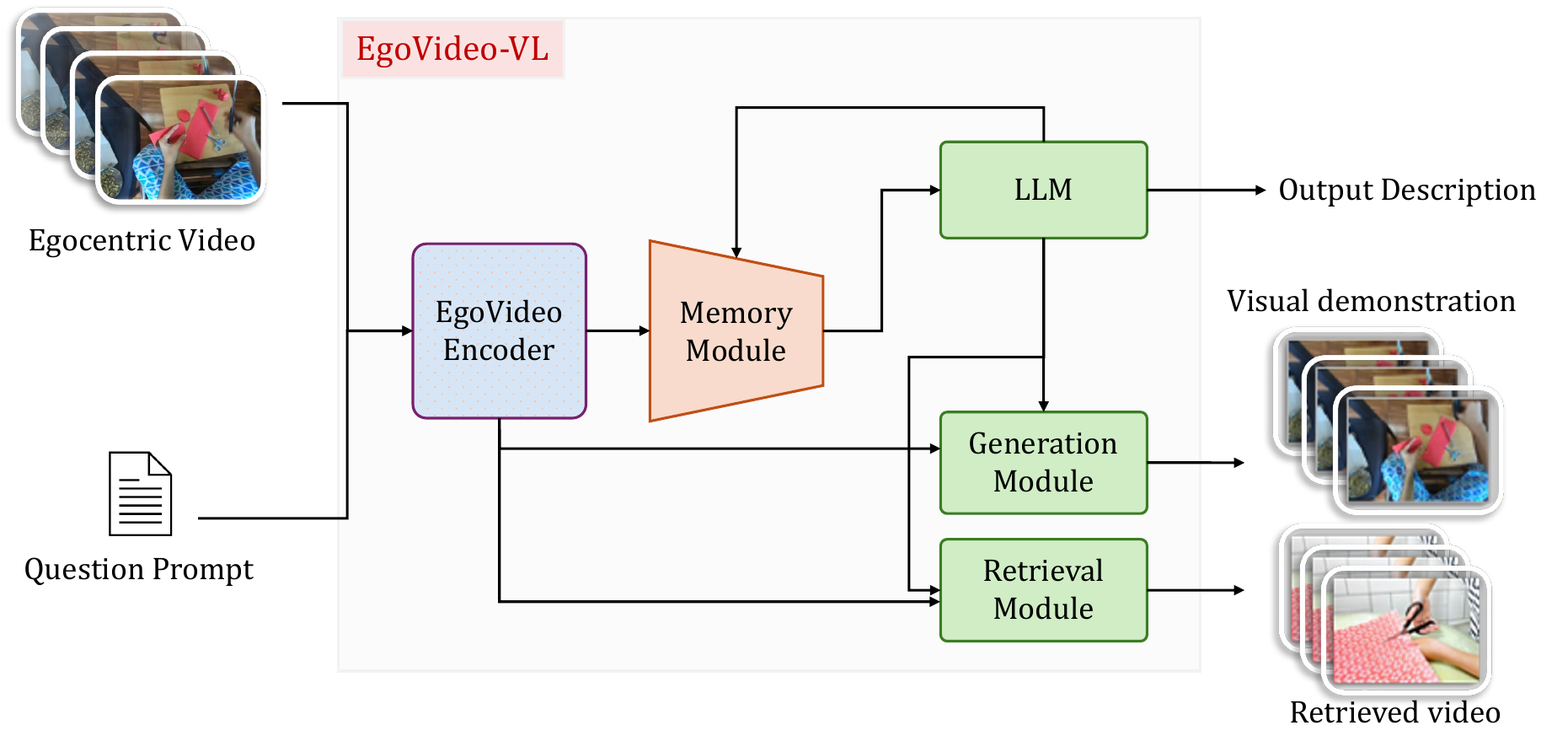}
    \caption{The overall structure of the EgoVideo-VL model. The visual encoder leverages the egocentric video foundation model, EgoVideo, while the LLM component utilizes InternLM. The memory module periodically processes video content, saving detailed descriptions and corresponding timestamps for historical context. The generation module predicts actions or creates visual demonstrations based on the current video frame and user prompts. The retrieval module retrieves third-person perspective videos, enabling users to watch and imitate skill-related tasks.}
    \label{fig:vl}
\end{figure*}

\section{Related Work}
\label{sec:related}
\subsection{Egocentric vision}
Egocentric vision focuses on analyzing videos captured from an actor-centered perspective, typically using wearable devices.
Research in this area has addressed various tasks, including action understanding~\cite{charadesego,huang2020mutual,zhang2024masked,huang2022compound}, pose estimation~\cite{wang2023scene,li2023ego}, hand-object interactions~\cite{chen2023gsdf,Shan20,zhang2022fine}, and scenes understanding~\cite{interaction-exploration, nagarajan2023egoenv}. Given the scarcity of high-quality training data, significant efforts have been made to develop large-scale egocentric datasets~\cite{ego4d,egobody,epickitchen,jia2020lemma,charadesego,ouyang2024actionvos,assembly101,finebio}. These datasets have enabled progress in egocentric AI by providing a diverse range of tasks and scenarios for model training and evaluation. Since egocentric videos are often in the form of long video streams, recent research has emphasized long video understanding from various perspectives. These include action recognition~\cite{min2021integrating,yang2023deco}, future prediction~\cite{furnari2019would,EgoProceLECCV2022}, video summarization~\cite{del2016summarization,xu2015gaze}, and memory mechanism~\cite{manigrasso2024online,plizzari2024spatial,wang2023memory}. 

While prior works have made remarkable advancements in understanding egocentric video and its specific tasks, Vinci distinguishes itself from prior works by introducing the first system built on an egocentric vision-language model (EgoVideo-VL). Unlike traditional approaches that address specific tasks or rely on offline processing, Vinci leverages the strong capabilities of VLMs to deliver a unified framework. It integrates real-time video and language processing, a memory module for temporal grounding, a generation module for creating actionable video demonstrations, and a retrieval module capable of finding how-to demos by others.

By combining the power of VLMs with egocentric vision, Vinci represents a significant step forward in the field. Its ability to process continuous video streams, understand historical and current contexts, and provide real-time assistance highlights its holistic and innovative approach to solving challenges in egocentric AI.

\subsection{Vision-language models}
Vision-language models have achieved significant progress in recent years, showcasing impressive capabilities in processing and understanding both visual and textual information. Open-source models like the LLaVA series~\cite{liu2023llava, liu2024llavanext}, the Qwen-VL series~\cite{QwenVL, Qwen2VL}, and the InternVL series~\cite{internvl, internvl2_5} have demonstrated state-of-the-art performance across a variety of vision-language tasks. However, these models are primarily trained on large-scale third-person datasets, making them less effective in the egocentric video domain.
With the introduction of the Ego4D \cite{ego4d} dataset, several studies have made strides toward addressing this gap. For instance, EgoVLP \cite{egovlp} utilizes contrastive learning for egocentric video and language pretraining, while EgoVLPv2 \cite{pramanick2023egovlpv2} enables cross-model fusion in video and language backbones. LaViLa \cite{lavila} refines the pre-training data using large language models to learn egocentric video and language representations. In addition, the champion solution of InternVideo-Ego4D~\cite{chen2022internvideo-ego4d} ensembles video-only and video-language features to significantly enhance the performance across multiple egocentric downstream tasks.

While these models have made significant advancements, they are not optimized for real-time applications or continuous interaction. Vinci's EgoVideo-VL bridges this gap by connecting the egocentric foundation model EgoVideo to a large language model (LLM), enabling free-form language interaction. This integration allows Vinci to process long egocentric video streams in real-time while leveraging the LLM for flexible, natural language-based communication. This capability distinguishes Vinci from existing models, making it uniquely suited for dynamic, real-world egocentric AI applications.

\begin{figure*}
    \centering
    \includegraphics[width=\linewidth]{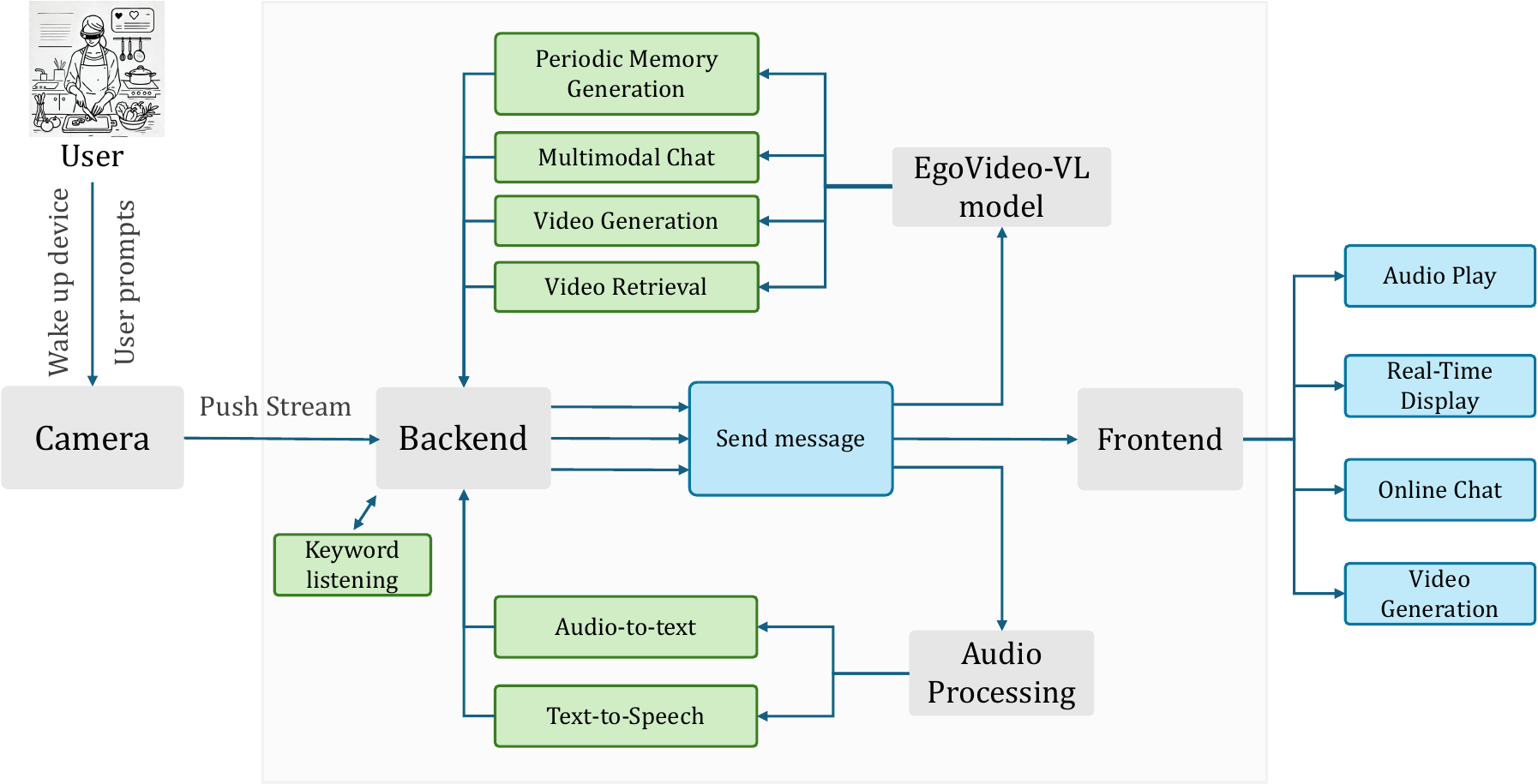}
    \caption{Overview of the Vinci system. The system integrates four components: the camera, frontend, backend, and models. The frontend is a web-based interface that displays the video stream and plays audio generated by TTS from the model's responses. The backend acts as the central hub, managing communication between the frontend, camera, and models. It listens for wake-up commands and, upon detection, activates the EgoVideo-VL model to process user prompts and deliver the corresponding outputs.}
    \label{fig:system}
\end{figure*}
\subsection{Streaming video understanding}
Conventional video understanding methods \cite{gao2020unsupervised, liu2022video,li2018jointly,song2021towards,lin2023univtg,chen2022dcan,yang2023basictad,chen2023elan,chen2024videomambasuite,chen2023videollm,zheng2023mrsn,huang2023weakly,chen2024cg-bench} typically use clips of videos or the entire video as input. While effective for offline analysis, this paradigm is not suited for AI systems that need to continuously gather and process information from dynamic environments. Streaming video understanding addresses this limitation by focusing on predictions based on current video frames, without relying on future information. This approach has been applied to tasks such as online action detection~\cite{Ghoddoosian2023WeaklySupervisedAS,Chen2022GateHUBGH,IIU}, action anticipation~\cite{9008264,8953649} and object tracking~\cite{8953931, SiamMask}.

Recently, advancements in vision-language models have paved the way for real-time interaction with users in streaming video contexts. For instance, Flash-VStream~\cite{zhang2024flashvstream} and VideoStreaming~\cite{qian2024videostreaming} both introduce memory mechanisms to process long video streams. VideoLLM-Online~\cite{VideoLLM-online} proposes a novel training objective that enables the model to determine the optimal moments to respond to user queries. Similarly, MMDuet~\cite{wang2024mmduet} refines the video-text interaction format to enhance temporal grounding. InternLM-XComposer2.5-OmniLive \cite{zhang2024internlmxcomposer25omnilive} integrates video and audio streaming processes with large language models to support multimodal interactions.

In procedural activity scenarios, prior work has shown potential as personalized assistants, leveraging large language models to provide textual guidance. However, textual instructions alone can be less intuitive for users. Vinci distinguishes itself by delivering comprehensive and user-friendly visual guidance. Its generation module creates customized how-to demonstrations, while its retrieval module accesses external how-to videos to provide complete procedural steps. This combination of visual and textual guidance enhances user experience, making Vinci an effective and versatile assistant for real-time streaming video applications, especially suited for egocentric scenarios.
\let\thefootnote\arabic{}
\renewcommand{\thefootnote}{\arabic{footnote}}

\section{Method}
Vinci comprises four main modules: 1) an input processing module that receives live-streaming video and audio while transcribing audio into text, 2) a vision-language model EgoVideo-VL that handles video understanding, language processing, and response generation, 3) a memory module for integrating and retrieving historical context, and 4) a generation module for creating instructional video demonstrations. These modules operate simultaneously for seamless real-time performance and robust user assistance.

\begin{figure*}
    \centering
    \includegraphics[width=\linewidth]{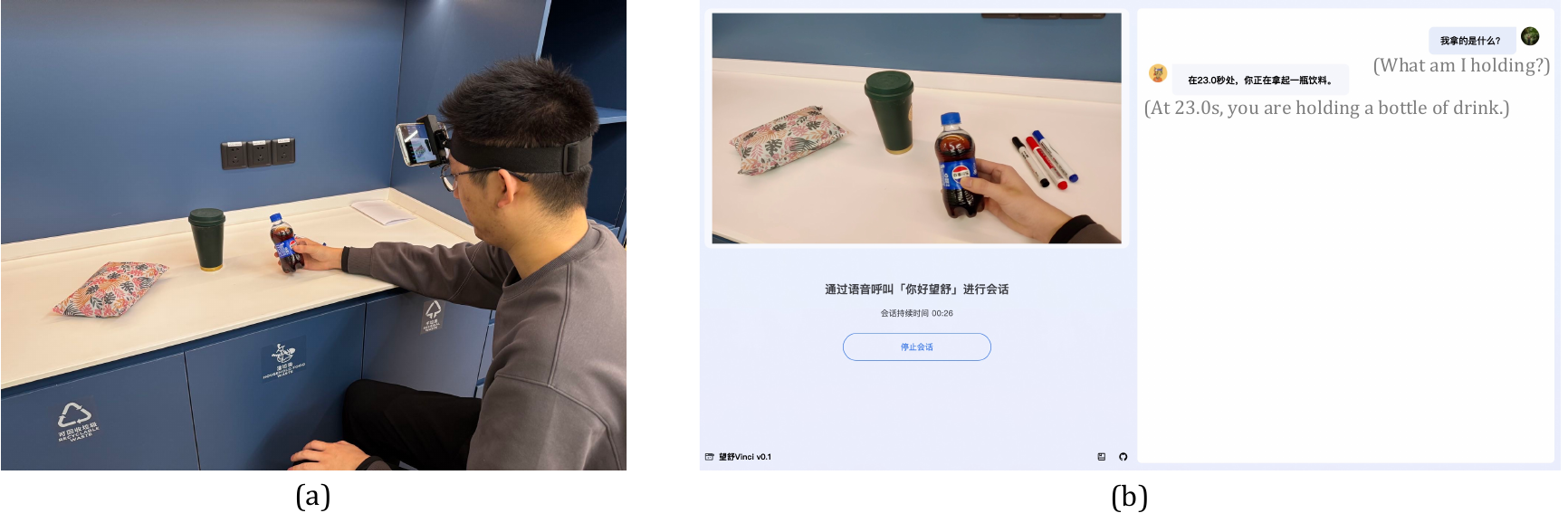}
    \caption{Real-world deployment of Vinci. (a) The deployed system on a OnePlus smartphone mounted on the user's head. (b) The web-based frontend. The displayed model output will also be played by audio for seamless interaction. }
    \label{fig:system2}
\end{figure*}

\subsection{Input processing}
The input processing module is designed to prepare live data streams from user devices for seamless downstream processing. It supports any device that uses the RTMP protocol for streaming, offering flexibility across various platforms such as smartphones and GoPro cameras. The module receives live-streaming video and audio inputs, with the audio transcribed into text using an automatic speech recognition (ASR) system. While we use Baidu’s ASR API for this purpose, the framework is adaptable to accommodate other ASR methods as needed.

In parallel, the video stream is processed to synchronize with the transcribed audio, maintaining alignment between visual and textual inputs. This capability ensures that all incoming data is structured and ready for real-time analysis by the vision-language model. 

\subsection{EgoVideo-VL}
At the core of Vinci’s capabilities is its vision-language model, EgoVideo-VL, which builds upon the state-of-the-art Egocentric foundation model EgoVideo~\cite{pei2024egovideo}. The overview of EgoVideo-VL is shown in Figure~\ref{fig:vl}. The visual token output from EgoVideo is connected to a large language model (LLM) to integrate visual and textual data. We employ instruction fine-tuning using a newly curated dataset to enhance EgoVideo’s ability to process egocentric video and align visual inputs with language queries. This dataset is constructed by combining data from three sources: Ego4D~\cite{ego4d}, EgoExoLearn~\cite{huang2024egoexolearn}, and Ego4D-Goalstep~\cite{song2024ego4d}. More specifically, we choose the narrations from Ego4D and EgoExoLearn, forming 4M video-text pairs. We then assign prompts to each video-text pair, based on LLM-refined prompts randomly chosen from a prompt template. For Ego4D-Goalstep, using the step and substep annotations, we generate question-answer pairs that focus on procedural task planning and historical reasoning.

We apply a two-stage fine-tuning strategy: in the first stage, we tune the model using Ego4D and EgoExoLearn to better align the vision and language tokens in the egocentric context. In the second stage, we incorporate all three datasets, with Ego4D-Goalstep\cite{song2024ego4d} data specifically designed to improve task planning and historical reasoning. The LLM component, InternLM-7B~\cite{cai2024internlm2}, remains fixed during training, providing a stable foundation for language understanding and generation. This carefully crafted training pipeline ensures that EgoVideo-VL is well-equipped to handle the diverse challenges of real-time egocentric AI applications.

\subsection{Memory module}
The memory module is a critical component of Vinci, designed to address the continuous and streaming nature of egocentric data. It operates by periodically capturing short video snapshots, describing the observed actions in detail, and storing the resulting textual descriptions along with their corresponding timestamps. This structured memory ensures that the system maintains a comprehensive record of past events. When a user interacts with Vinci, the memory module provides relevant historical context to EgoVideo-VL, enabling the system to temporally ground past actions, summarize user activities, and answer queries requiring historical reasoning. By integrating this memory framework, Vinci achieves a robust understanding of temporal sequences, supporting advanced functionalities like multi-step task planning and natural language grounding.

\subsection{Generation module}
The generation module in Vinci is responsible for creating visual how-to demonstrations, enabling users to understand and perform tasks requiring detailed visual guidance. It is built upon SEINE~\cite{seine}, an image-to-video generation model from the open-source video generation system Vchitect~\cite{lavie,seine,latte,vbench,vbench++++}, which has been fine-tuned to address the specific challenges of egocentric video generation.

To fine-tune SEINE, we curate a specialized dataset filtered from our instruction tuning dataset using two key criteria: (1) videos must exhibit smooth global motion, achieved by selecting only those with a maximum optical flow below a defined threshold, and (2) the verbs in the associated text must occur with reasonable frequency to prevent the model from collapsing during training due to rare verbs.

For video generation, EgoVideo-VL first determines whether generation is necessary based on the user’s query. When required, it provides the user query and the most recent video frame to the generation module, which then outputs a 2-second video illustrating the requested action. This functionality equips Vinci with the capability to deliver actionable, visual guidance tailored to user needs.

\subsection{Retrieval Module}
The retrieval module in Vinci aims to automatically retrieve how-to videos given the user query, providing visual demonstrations and references for users to complete the ongoing tasks. Different from the Generation Module that directly generates the visual guidance via a video diffusion model, the videos are retrieved from a large video database containing instructional videos in HowTo100M~\cite{howto100m}.

To achieve this goal, we adopt the readily available retrieval module in EgoInstructor~\cite{xu2024retrieval}, a model that is trained on pseudo paired egocentric-exocentric videos~\cite{ego4d,howto100m}. Given a user query in the text format (\emph{e.g.} show me how to cut a tomato), the model first extracts text features and computes normalized cosine similarity with cached video features stored in the database. Then, it selects the top-K videos with the largest cosine similarity using FAISS~\footnote{\url{https://github.com/facebookresearch/faiss}}, and we choose K=3 in our system. This functionality enables Vinci to provide visual guidance sourced from a wide range of online instructional videos, reducing the need for manual search on Google or YouTube.

\subsection{System integration}
The Vinci system is a comprehensive integration of four major components: the camera, frontend, backend, and models, all working together to deliver real-time egocentric AI assistance. The overview workflow can be seen in Figure~\ref{fig:system}.

The camera serves as the input source, recording live streaming video and audio that are transmitted to the backend using the RTMP protocol. Vinci supports various camera setups, including: 1) Smartphones equipped with streaming apps, 2) Wearable cameras such as GoPro or DJI cameras, and 3) Standard webcams configured with FFmpeg for streaming.

The backend acts as the central hub for processing and communication. It hosts HTTP services to receive and manage live streams, detect wake-up keywords in the audio input, and handle interactions with the models. For the detection of wake-up keywords, we apply an API-based solution \footnote{\url{https://www.chumenwenwen.com/en}}. Once the EgoVideo-VL model generates a response, the backend processes the output and pushes both the live stream and the response back to the frontend.

\begin{figure*}[h]
    \centering
    \includegraphics[width=\linewidth]{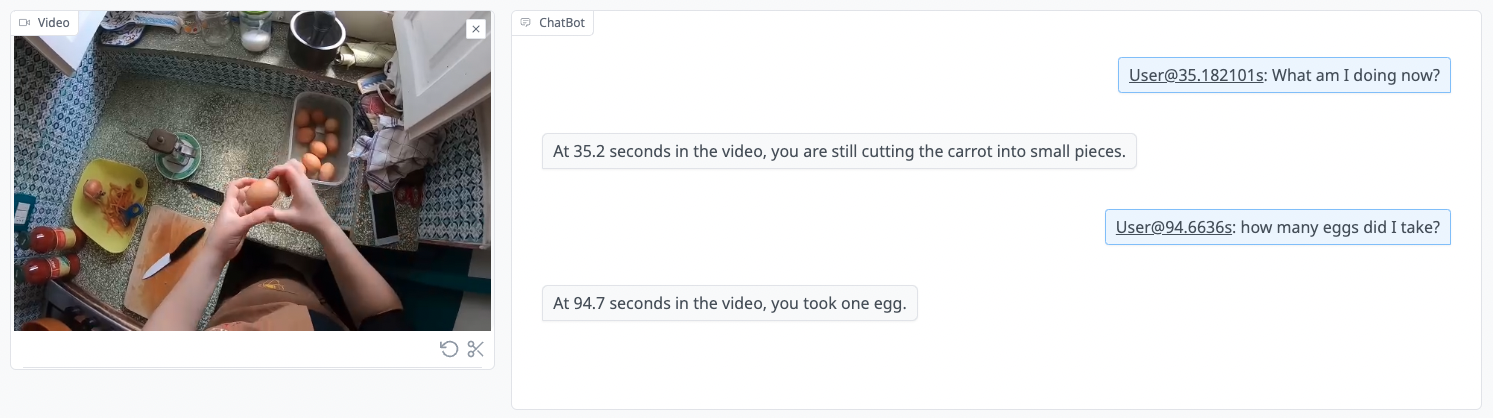}
    \caption{Example of Vinci’s ability to analyze the current video state and accurately respond to user queries. In this scenario, at 35.2 seconds, Vinci correctly identifies the ongoing action as cutting carrots. At 94.7 seconds, Vinci accurately recognizes the current scene, confirming that only one egg is being held in hand.}
    \label{fig:current5}
\end{figure*}

\begin{figure*}[h]
    \centering
    \includegraphics[width=\linewidth]{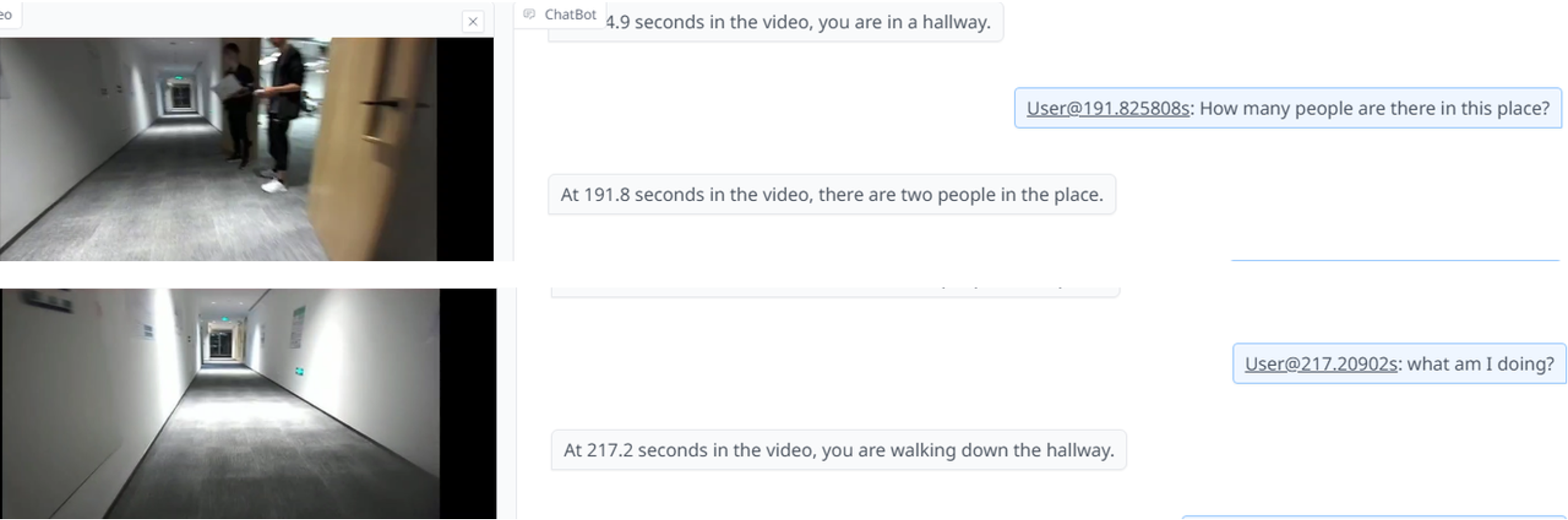}
    \caption{Example of Vinci’s ability to generalize to embodied navigation scenarios without additional tuning. In this case, Vinci accurately identifies the agent walking down the hallway at 217.2 seconds and correctly counts the number of people present at 191.8 seconds, demonstrating its versatility in understanding diverse video contexts.}
    \label{fig:current2}
\end{figure*}

\begin{figure*}[h]
    \centering
    \includegraphics[width=\linewidth]{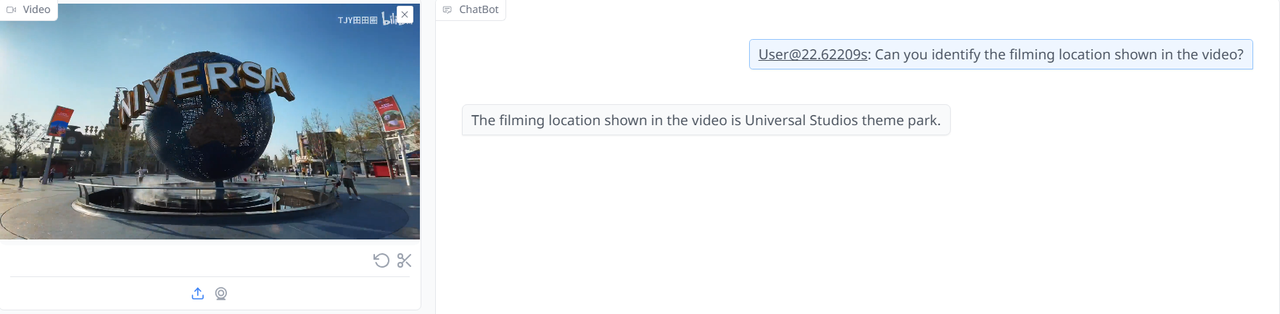}
    \caption{Example of Vinci’s general video scene understanding capabilities. In this scenario, Vinci successfully identifies the filming location as Universal Studios theme park, demonstrating its general video scene understanding abilities.}
    \label{fig:current1}
\end{figure*}

\begin{figure*}[h]
    \centering
    \includegraphics[width=\linewidth]{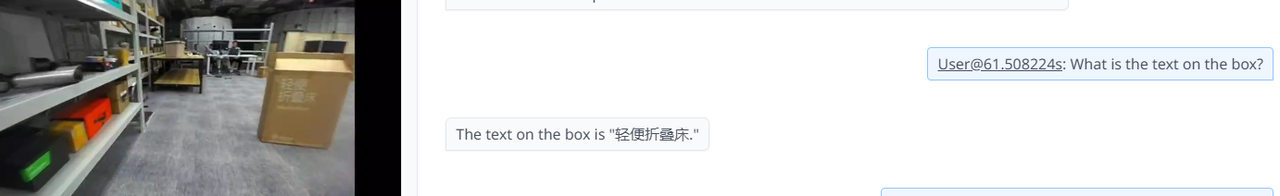}
    \caption{Example of Vinci’s strong OCR capability. In this scenario, Vinci correctly recognizes the Chinese characters on the box when prompted by the user, showcasing its ability to accurately process and interpret text within the video.}
    \label{fig:current4}
\end{figure*}

The frontend is a JavaScript-based web application that provides a user-friendly interface. It displays the live video stream and synchronously plays the accompanying audio. Additionally, it uses a text-to-speech (TTS) system to convert the text responses from EgoVideo-VL into audio, ensuring seamless interaction with the user.

The models, including EgoVideo-VL, the retrieval module and the generation module, connect directly with the backend. They receive input in the form of video frames and transcribed text from the backend. The models process the input to generate either textual responses or short video clips, which are then sent back to the backend for further delivery to the frontend.

This tightly integrated system ensures a smooth flow of data and functionality between the components, enabling Vinci to deliver robust real-time assistance in a variety of scenarios. We open-source the full system including the frontend, backend, and models. An example of the deployed Vinci system is shown in Figure~\ref{fig:system2}. We provide two options for experiencing our system. Users can deploy our system and test the interaction using wearable cameras or smartphones, or they can use a Gradio-based system to test our system with uploaded videos.

\section{Experiments}
\subsection{Qualitative analysis}
In this section, we present a qualitative analysis of Vinci’s capabilities using examples from its Gradio demo with uploaded long videos. Readers can refer to our project webpage for a complete demonstration video showcasing the system in action. This analysis highlights Vinci’s ability to perform a wide range of tasks, including current scene understanding, history temporal grounding, video summarization, future planning, action prediction, and video generation. Below, we provide an overview of each capability, along with illustrative examples.

\subsection{Current scene undersanding}
Vinci exhibits a robust ability to analyze the current scene and provide detailed descriptions based on live video input and user queries. Figures~\ref{fig:current5}, \ref{fig:current2}, \ref{fig:current1}, and \ref{fig:current4} illustrate these capabilities through various examples.

In Figure~\ref{fig:current5}, Vinci accurately recognizes the user's current action at 35.2 seconds when prompted ``what am I doing now?", and correctly counts the number of eggs in hand at 94.7 seconds upon request. In Figure~\ref{fig:current2}, we directly use Vinci on a video of embodied agent navigation without tuning. In the figure, Vinci can correctly recognize the current scene and count the number of people objects. This shows that Vinci also has strong potential in Embodied AI.

Figure~\ref{fig:current1} highlights Vinci's general video understanding capability, where it identifies the location as Universal Studios. Additionally, Figure~\ref{fig:current4} demonstrates Vinci's OCR skills, correctly recognizing Chinese characters displayed on a box.

\subsection{Temporal grounding}
\begin{figure*}
    \centering
    \includegraphics[width=\linewidth]{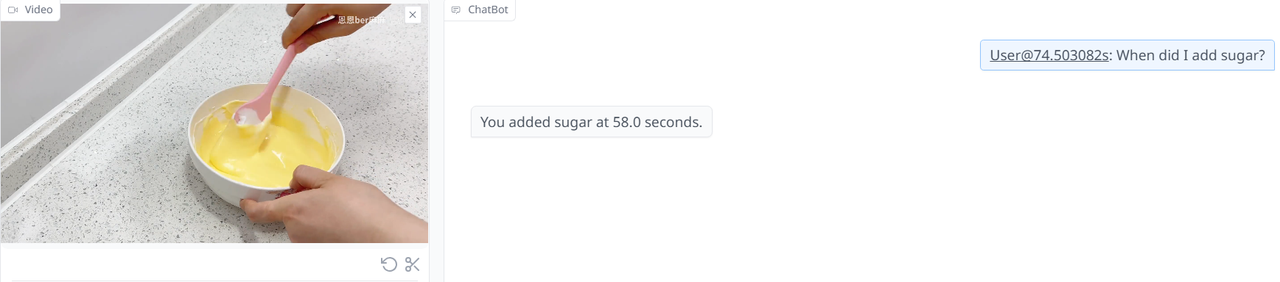}
    \caption{Vinci can perform temporal grounding, helping the user to locate the timestamp of specific previous actions.}
    \label{fig:grounding1}
\end{figure*}

\begin{figure*}
    \centering
    \includegraphics[width=\linewidth]{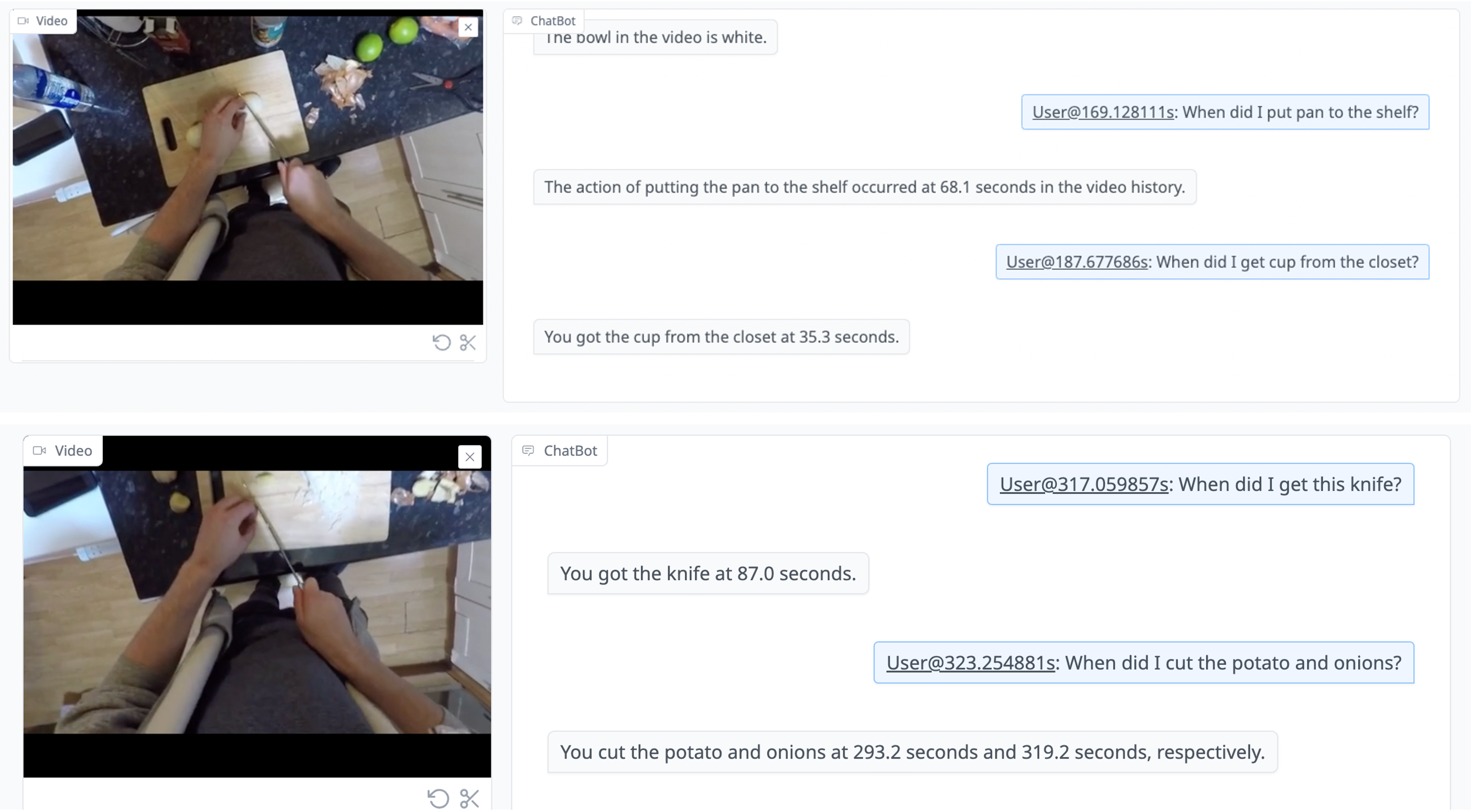}
    \caption{Vinci can correctly locate previous actions even when they are queried out of chronological order or with a significant time gap between the query and the actual event. Vinci can also ground multiple actions simultaneously.}
    \label{fig:grounding2}
\end{figure*}

Leveraging its memory module, Vinci demonstrates the ability to retrieve and reference past actions or events in response to user queries, showcasing robust temporal grounding capabilities. In Figure~\ref{fig:grounding1}, the user asks Vinci ``When did I add sugar" at 74s. Vinci accurately responds that the sugar was added at 58 seconds, effectively pinpointing the exact moment of the action. Similarly, Figure~\ref{fig:grounding2} highlights Vinci's ability to ground previously occurred actions, even when they are queried out of chronological order or with a significant time gap between the query and the actual event. In the bottom part of Figure~\ref{fig:grounding2}, Vinci also shows the ability to ground multiple actions simultaneously.

\subsection{Video summarization}
\begin{figure*}[]
    \centering
    \includegraphics[width=\linewidth]{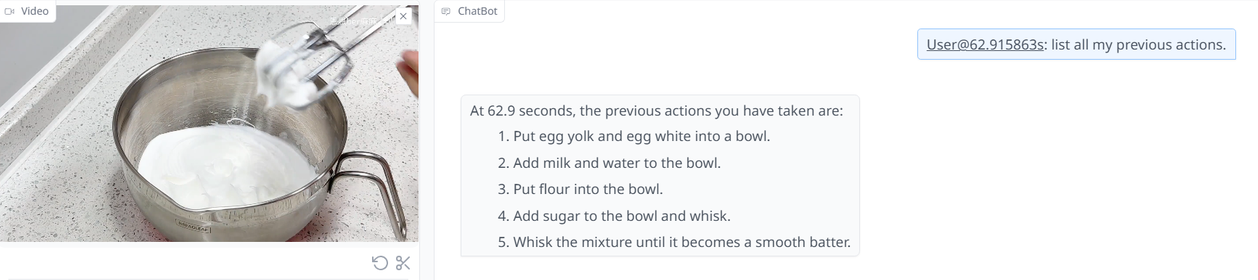}
    \caption{Vinci can make the stepwise summarization of long-horizon videos.}
    \label{fig:summarize}
\end{figure*}

Vinci demonstrates strong video summarization capabilities, providing concise overviews of complex, multi-step activities.
In Figure~\ref{fig:summarize}, when prompted at 62.9 seconds, Vinci effectively summarizes the key actions performed up to that point. It identifies a series of cooking steps, including separating eggs, adding various ingredients to a bowl, and whisking the mixture until smooth.
This example highlights Vinci’s ability to distill long sequences of actions into coherent summaries, offering users a quick and clear understanding of prior events without needing to review the entire video.

\begin{figure*}
    \centering
    \includegraphics[width=\linewidth]{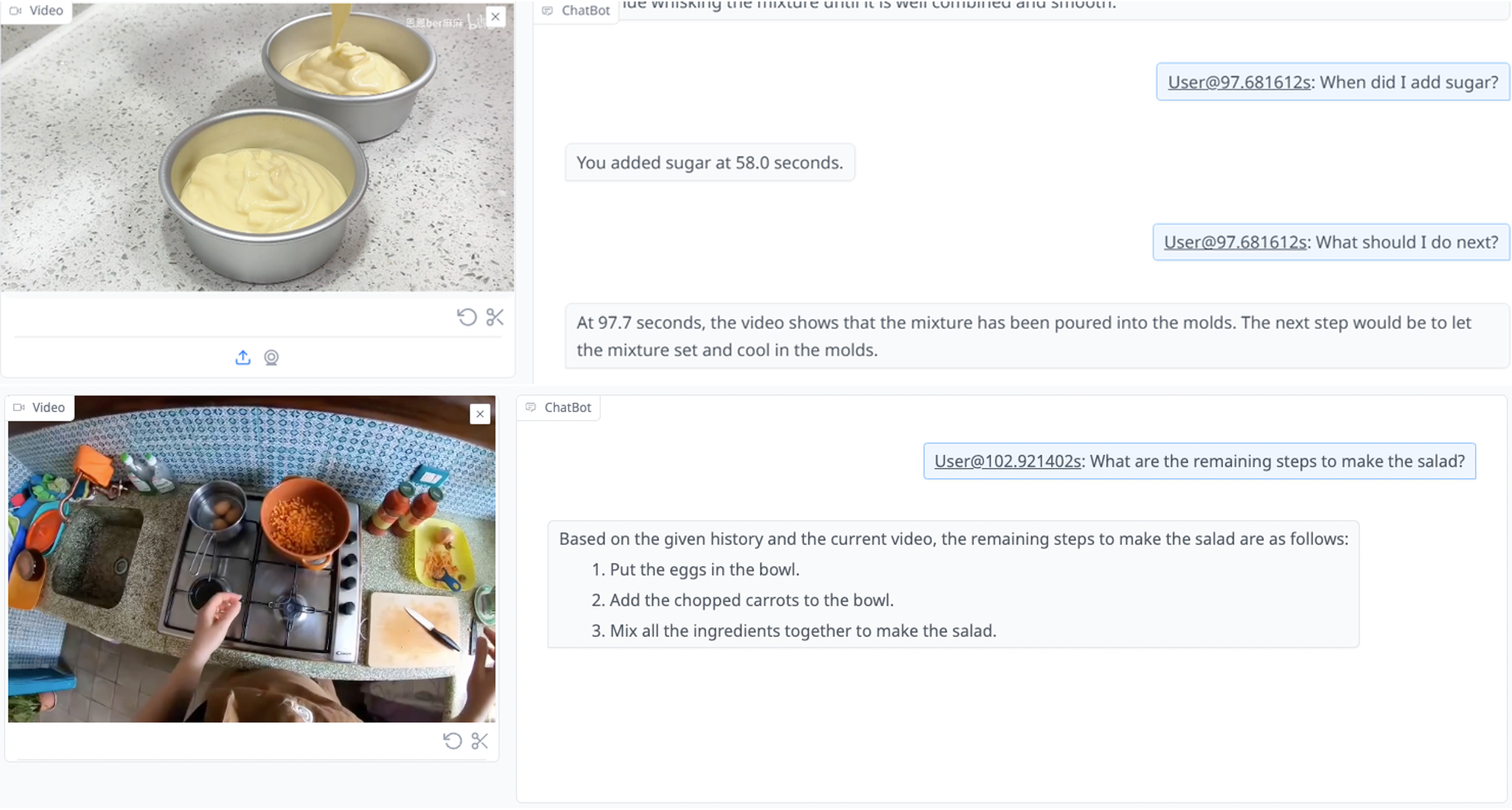}
    \caption{Vinci can provide future planning based on the current video state. In the bottom example, Vinci demonstrates its ability to create a detailed multi-step plan based on the history and the current video.}
    \label{fig:future1}
\end{figure*}


\subsection{Future planning}
Vinci excels in planning future steps or predicting subsequent actions by leveraging historical context and real-time observations.
In Figure~\ref{fig:future1}, Vinci analyzes the current video state, recognizing that the mixture has already been poured into molds. It then suggests the logical next step—allowing the mixture to set and cool in the molds.
In the bottom part of Figure~\ref{fig:future1}, Vinci demonstrates its ability to create a detailed multi-step plan. When prompted about the remaining steps to complete a salad, it synthesizes historical and current context to outline the necessary actions, such as adding specific ingredients and mixing them together.
These examples showcase Vinci’s practical ability to provide guidance, with TTS techniques, Vinci can be a practical tool for step-by-step task planning and procedural reasoning.

\subsection{Cross-view video retrieval}
\begin{figure*}[]
    \centering
    \includegraphics[width=\linewidth]{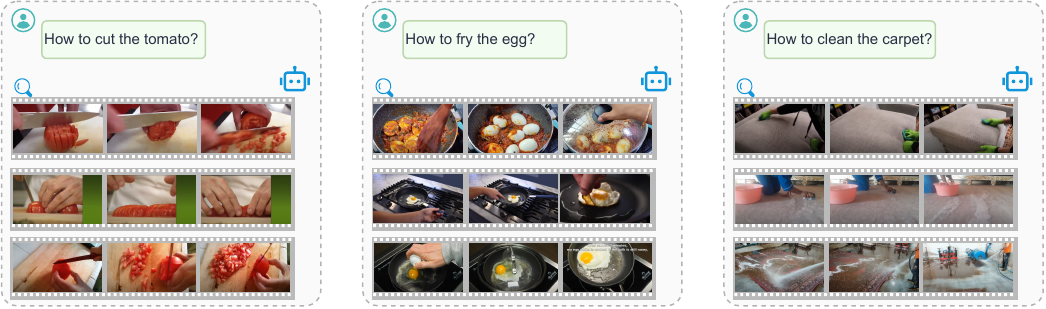}
    \caption{Examples of retrieval results from Vinci's retrieval module. The system identifies relevant how-to demonstration videos from a database (e.g., HowTo100M dataset) based on the user's query. The retrieved videos provide clear, third-person visual guidance, such as different methods and tools for cutting a tomato, helping users perform tasks with confidence and precision.}
    \label{fig:retrieval}
\end{figure*}
With a retrieval module, Vinci can also find how-to demos from the database (HowTo100M dataset~\cite{howto100m}), so that the users can learn how others perform the same task. In Figure~\ref{fig:retrieval} we show several examples of the retrieval result.  In Figure~\ref{fig:retrieval}, we illustrate several examples of the retrieval results. For instance, when a user queries ``How to cut the tomato," Vinci retrieves multiple instructional videos demonstrating the cutting process from various angles and with different tools. This functionality bridges the gap between egocentric and third-person perspectives, offering users diverse and detailed visual guidance to complete their tasks effectively.

\begin{figure*}[htbp!]
    \centering
    \includegraphics[width=\linewidth]{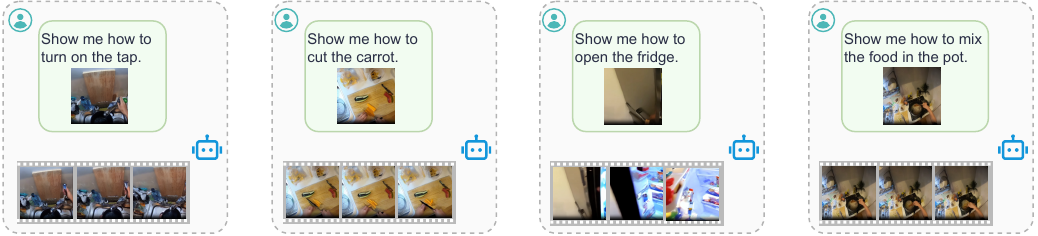}
    \caption{Examples of visual demonstrations generated by Vinci's generation module. Each 2-second video clip is synthesized based on the latest video frame and a user prompt, showcasing the predicted action to assist in task completion. The module is fine-tuned on egocentric video data, ensuring high-quality outputs for egocentric scenarios.}
    \label{fig:generation}
\end{figure*}
\subsection{Action prediction}
Vinci leverages its generation module to provide visual demonstrations, aiding users in task completion. By combining the latest video frame with the user prompt, Vinci generates a predicted action as a 2-second video clip. Figure~\ref{fig:generation} illustrates several examples of these generated outputs. The generation module, fine-tuned on egocentric video data, performs optimally for egocentric scenarios but is less effective for third-person video generation tasks.

\section{Potential Applications}
Vinci, with its real-time processing capabilities, robust scene understanding, and ability to generate visual guidance, opens up several exciting possibilities for applications in both Augmented Reality (AR) / Virtual Reality (VR) environments and the broader field of Embodied AI.

\paragraph{AR/VR Assistants} In AR/VR environments, Vinci can serve as a powerful, hands-free assistant, providing real-time assistance to users as they interact with digital or physical worlds. By integrating Vinci into AR glasses, headsets, or other wearable devices, users can access contextual information through natural language interaction and receive step-by-step guidance without needing to break immersion or look away from the task at hand.

For example, Vinci can provide real-time instructions during a complex AR training simulation, helping users by recognizing their actions and offering timely feedback. Whether in medical training, industrial design, or remote support for technical troubleshooting, Vinci can guide users through intricate processes with real-time, dynamic visual and textual cues. The combination of scene understanding, task planning, and video generation features makes Vinci an ideal candidate for tasks requiring ongoing decision-making and situational awareness, such as emergency response training or complex machine maintenance.

\paragraph{Embodied AI} Vinci’s capacity to understand and interact with its environment through real-time video input also positions it as a valuable tool for embodied AI applications. Embodied AI systems, which involve agents that perceive and act in the physical world, can benefit significantly from Vinci's ability to provide both high-level understanding and low-level task execution guidance. Vinci can be integrated into robots or autonomous agents to enable real-time scene analysis and interaction with objects, people, or other agents in dynamic environments.

For example, in industrial robotics, Vinci can guide robots through assembly lines, detecting and responding to changes in the environment, and even predicting future actions based on ongoing tasks. Similarly, in healthcare, embodied agents equipped with Vinci could assist elderly individuals by recognizing daily activities, offering reminders, and even providing real-time visual instructions for tasks such as meal preparation or personal care.

Moreover, Vinci’s ability to integrate both short- and long-term contextual understanding makes it an ideal solution for navigating complex environments where historical context is critical, such as in autonomous navigation for drones or ground robots. By combining temporal grounding, video summarization, and task prediction, Vinci enhances the adaptability and functionality of embodied systems, allowing them to make informed decisions in real-time while drawing from past experiences.

\paragraph{Multi-Domain support and future prospects} As Vinci continues to evolve, its potential applications extend across various domains that require real-time, context-aware interaction. In combination with AR/VR environments and embodied AI, Vinci could be adapted for use in smart homes, educational settings, and advanced human-robot collaborations, where it could provide both procedural support and autonomous decision-making capabilities. With further improvements, Vinci could enable more complex, seamless interactions in mixed-reality environments, creating more immersive and intelligent virtual assistants, while also contributing to the broader vision of autonomous, intelligent robots capable of interacting with their environment in a natural and intuitive way.
\section{Conclusion}
In this technical report, we introduced Vinci, a real-time embodied smart assistant powered by the egocentric vision-language model, EgoVideo-VL. Vinci represents a significant advancement in the field of egocentric AI by integrating state-of-the-art vision-language processing with real-time streaming capabilities. It combines multiple modules, including input processing, memory management, video retrieval and generation, to provide seamless interaction, robust historical reasoning, and detailed visual guidance.

We demonstrated Vinci’s capabilities across various tasks, such as current scene understanding, temporal grounding, video summarization, future planning, and procedural video generation. These experiments highlight Vinci's ability to process and respond to complex user queries in dynamic environments, providing a hands-free, intuitive user experience.

A key contribution of Vinci is the novel instruction-tuning dataset, curated from Ego4D, Ego4D-Goalstep, and EgoExoLearn datasets. This dataset enabled the fine-tuning of EgoVideo-VL, equipping it with the ability to process egocentric video data effectively and interact with users in free-form natural language. Moreover, by fine-tuning SEINE, Vinci extends its utility with high-quality visual how-to demonstrations, addressing a significant limitation of previous systems.

To foster future research and innovation, we have released the model parameters and the full deployment source code, including both the web frontend and backend. We hope Vinci can serve as a foundation for further explorations in egocentric vision, vision-language models, and embodied AI applications. 
{
    \small
    \bibliographystyle{ieeenat_fullname}
    \bibliography{main}
}


\end{document}